\DocumentMetadata{
	lang=en,
	pdfversion = 2.0,
	pdfstandard = ua-2, 
	tagging=on,
	tagging-setup={math/setup=mathml-SE}
}

\documentclass[12pt]{amsart}

\usepackage{amsmath,amssymb}
\usepackage[T1]{fontenc}
\usepackage[utf8]{inputenc}
\usepackage{microtype}
\usepackage{hyperref}
\hypersetup{
  colorlinks=true,
  linkcolor=blue!60!black,
  citecolor=blue!60!black,
  urlcolor=blue!60!black
}
\usepackage{enumitem}
\usepackage{booktabs}
\usepackage{graphicx}
\usepackage{xcolor}
\usepackage{geometry}
\usepackage{hyphenat}
\newcommand{\phaseprompt}[2]{%
  \par\addvspace{4pt}%
  \noindent\textbf{#1}\par\nobreak
  \noindent #2\par
}

\title[A Six-Phase AI Workflow for Interactive Visualizations]{Leveraging Generative AI to Design Accessible Interactive
       Visualizations for Undergraduate Mathematics:
       A Six-Phase Workflow}

\author{Mahesh Sunkula \and Kuan-Hua (Joseph) Chen}

\address{Department of Mathematics, Purdue University}
\email{msunkula@purdue.edu \quad jchen@purdue.edu}



\begin{document}

\begin{abstract}
Interactive visualizations support conceptual understanding in
undergraduate mathematics, but building them has required programming
expertise most instructors lack. Using a design-based research approach,
we develop, deploy, and evaluate a six-phase workflow (Foundation,
Customization, Mathematical Depth, Application, Accessibility,
Pedagogical Control) that uses generative AI to build WCAG~2.2 Level~AA
compliant visualizations without programming. The six phases structure
every prompt, scaffold the AI's code generation, and define where human
verification is applied. We ask whether the structure reliably yields
correct and accessible tools, whether it runs both backward
(reverse-engineering prompts from a finished tool) and forward (generating
a tool from a plain-language idea), and what verification each phase
requires. Across four deployed tools spanning calculus, multivariable
calculus, and differential equations, we evaluate mathematical
correctness against closed forms, accessibility through automated and
manual screen-reader testing, and the errors that recurred. The structure
produces structurally complete first-pass tools, but human verification
remains mandatory at every phase: each output must be checked for
mathematical correctness, accessibility, and pedagogical fit before the
next phase begins. The workflow is platform-independent and serves both
instructors and students.
\end{abstract}

\maketitle

\section{Introduction}

As mathematics instructors, we face a persistent gap between what we can
visualize mentally and what we can communicate to students. The gradient
perpendicular to level curves, a saddle point's instability, the way a
linear map shears a parallelogram: these are vivid to us and opaque on
a whiteboard. Interactive visualizations close that gap: students
manipulate parameters in real time and build the geometric intuition that
supports their analytical work. The problem has always been that creating
such tools requires programming expertise most of us do not have and time
none of us can spare.

Generative AI changes this. We can now describe what we want in the same
language we use with a colleague and receive working code. The key, we
found, is structure. A single large prompt produces mediocre results.
Breaking the task into six focused phases, each with a clear
deliverable validated before the next begins, produces tools that
are mathematically correct, pedagogically purposeful, and fully
accessible. This article documents that six-phase workflow and the two
modes in which we use it.

\subsection{The pedagogical challenge}

The visualization problem extends across the undergraduate curriculum.
In differential equations, students who can write down a solution formula
often cannot describe what it looks like. In linear algebra, matrix
transformations become meaningful when students watch a map stretch and
shear a region in real time. The research literature supports what we
observe: students who engage with interactive visualizations develop
deeper conceptual understanding than those who work exclusively with
symbolic representations~\cite{Tall1991,Presmeg2006,Habre2001}. Duval's
theory of semiotic representation argues that mathematical understanding
requires coordination between symbolic, graphical, numerical, and verbal
registers~\cite{Duval2006}; interactive tools support exactly that
coordination.

Not all visualizations help, however: poorly designed tools can introduce
extraneous cognitive load and hinder rather than support understanding.
Pedagogically principled design remains essential even when AI handles the
implementation.

\subsection{Research questions and approach}

We treat the workflow as the object of a design-based research (DBR)
study. DBR develops and refines an intervention through iterative cycles
of design, deployment, and analysis in authentic instructional settings,
producing both a usable artifact and transferable design
principles~\cite{DBRC2003,AndersonShattuck2012}. That is exactly the
situation here: each tool was built, deployed in a real course context,
and revised across the six phases, and each cycle surfaced design
principles about where the AI succeeds and where human expertise remains
indispensable. Our unit of analysis is the workflow itself, evaluated
through the tools it produces.

We address three research questions:
\begin{itemize}[leftmargin=1.5em]
\item \textbf{RQ1.} Does the six-phase structure reliably yield
  visualizations that are mathematically correct and WCAG~2.2 Level~AA
  compliant?
\item \textbf{RQ2.} Does the workflow operate in both directions
  (backward, reverse-engineering a phase-aligned prompt set from a
  finished tool, and forward, generating a tool from a plain-language
  pedagogical idea), and is it independent of the particular AI platform?
\item \textbf{RQ3.} What human verification does each phase still require,
  and what categories of error recur?
\end{itemize}

To answer them we provide explicit phase-by-phase prompts from four
deployed tools, an account of the two-directional pipeline with
worked examples, and an evaluation of mathematical correctness,
accessibility compliance, and pedagogical fit, organized phase by phase.
The contribution is both practical and methodological: instructors gain
adaptable prompts and a verification protocol, while the field gains a set
of design principles for AI-assisted educational tool creation. Our aim is
that a reader finishes ready to describe a first visualization idea to an
AI system and use the result, with a clear-eyed sense of what must be
checked before it reaches students.

\section{Theoretical Background}

\subsection{Visualization in mathematics learning}

Tall and Vinner's~\cite{TallVinner1981} distinction between concept
definition and concept image illuminates why visualization matters:
students frequently build concept images that conflict with formal
definitions, particularly in multivariate calculus, where geometric
intuition is hard to develop from symbolic manipulation alone. Dynamic
visualizations enrich concept images by providing varied examples and
making abstract relationships visible~\cite{Tall1991}. Empirical support
is strong: Mart\'{\i}nez-Planell and Trigueros~\cite{MartinezPlanell2012}
found that coordinating algebraic and geometric representations of
two-variable functions is strongly correlated with conceptual
understanding, and that dynamic software facilitates this coordination.
Weber and Thompson~\cite{Weber2014} similarly document how students'
images of two-variable functions and their graphs shape, and often
constrain, the conceptual understanding they are able to build.

\subsection{Active learning and student-created representations}

Constructivist theory holds that learners build understanding through
active engagement rather than passive reception~\cite{Piaget1970,
vonGlasersfeld1995}. When students create visualizations, they must
articulate mathematical relationships precisely enough for the AI to
implement them, a form of mathematical communication that surfaces
implicit gaps and misconceptions~\cite{Kaput1987}. The productive
struggle involved in this process is itself pedagogically
valuable~\cite{HiebertGrouws2007}. The capacity to describe a
mathematical relationship precisely enough for a machine to act on it
also connects to broader efforts to define computational thinking for
mathematics and science classrooms~\cite{Weintrop2016}.

\subsection{Accessibility and Universal Design for Learning}

Universal Design for Learning principles advocate for accessible
materials from the outset~\cite{RoseMeyer2002,CAST2018}. Most
web-based mathematics materials fail basic accessibility
standards~\cite{Alajarmeh2020}, and mathematics accessibility requires
particular attention to conveying relationships through non-visual
channels~\cite{KarshmerBledsoe2002}. WCAG~2.2~\cite{W3C2023} provides
concrete success criteria. This workflow enables instructors to specify
accessibility requirements in natural language, with AI implementing the
technical details, though human testing with assistive technology users
remains essential~\cite{ShinoharaWobbrock2011}.

\subsection{Generative AI in education}

Large language models offer real opportunities for educational
content generation, with important caveats about accuracy and
overreliance~\cite{Kasneci2023}. Code generation research demonstrates
high success rates for well-specified tasks but reveals that generated
code may contain subtle bugs or fail on edge cases~\cite{Chen2021,
Liang2023}. Specificity, iterative refinement, and domain knowledge
significantly improve output quality~\cite{Cooper2023}. Our six-phase
structure is a direct response to these findings: it maximizes AI
effectiveness while preserving human oversight at every stage.

\section{The Six-Phase Workflow}

The six-phase workflow addresses a fundamental challenge in AI-assisted
development: giving an AI system a single complex prompt produces
mediocre results. Breaking the same task into six targeted phases, each
with a clear deliverable and a validation step, consistently produces
tools that are mathematically correct, pedagogically purposeful, and
fully accessible. The phases are constant whether you are prompting
interactively (Stage~1) or asking the AI to generate all six prompts
from a pedagogical idea (Stage~2).

\subsection{Phase structure and rationale}

\medskip
\noindent\textbf{Phase~1: Mathematical Foundation and Core Interactivity.}
Establish the fundamental mathematical concept and basic user
interaction. Focus on correctness of mathematical implementation without
additional features that could complicate debugging. The deliverable is
a working visualization that accurately represents the core mathematical
relationship, even if pedagogically incomplete.

\medskip
\noindent\textbf{Phase~2: Customization and User Control.}
Add mechanisms for users to modify parameters, select from preset
examples, or input custom functions. This phase transforms a
demonstration into an exploration tool. Validation logic for user
inputs (domain restrictions, syntax checking) is implemented here to
prevent downstream errors.

\medskip
\noindent\textbf{Phase~3: Mathematical Depth and Connections.}
Incorporate related mathematical concepts that deepen understanding of
the core idea: derivative information, related geometric objects (tangent
planes, normal vectors), or abstract relationships (orthogonality,
optimization directions). The goal is to connect isolated concepts into a
coherent mathematical framework.

\medskip
\noindent\textbf{Phase~4: Applications and Algorithmic Connections.}
Bridge abstract mathematics to applications or computational methods:
optimization algorithms, numerical integration, physical simulations.
This phase makes concrete the student question ``when would we use
this?''

\medskip
\noindent\textbf{Phase~5: Accessibility and Inclusive Design.}
Implement WCAG~2.2 Level~AA compliance: keyboard navigation with visible
focus indicators (SC~2.4.7), minimum touch target sizes of $44\times 44$
CSS pixels (SC~2.5.8), ARIA live regions for dynamic content
announcements, skip-to-content links, and \texttt{prefers-reduced-motion}
support. Accessibility is addressed here rather than retrospectively, but
it should never be treated as optional.

\medskip
\noindent\textbf{Phase~6: Pedagogical Control and Scaffolding.}
Add layer management or progressive disclosure: checkboxes or toggles
that let instructors build up complexity incrementally during a lecture.
Reset functionality allows students to experiment freely. This phase
transforms a tool that displays mathematics into one that teaches it.

\subsection{Tool selection: Claude and Gemini}

We used both Anthropic's Claude and Google's Gemini. Each has strengths:
Claude excels at React-based component architecture, accessibility
implementation, and Three.js 3D graphics; Gemini is particularly strong
with Plotly.js visualizations and HTML5 Canvas animations. Both produce
high-quality results when prompts follow the six-phase structure. The
technology stack matters more than the platform. This holds in Stage~2 as
well: both Claude and Gemini can generate the full six-phase prompt set
from a plain-language description of a pedagogical idea and then build the
tool from it. We deliberately carried the two Stage~2 examples
end-to-end on different platforms, Example~3 entirely with Gemini and
Example~4 entirely with Claude and Claude Code, to show that the phase
structure, not the choice of model, is what carries the work (see
\S\ref{sec:pipeline}).

\section{Interactive Visualizations}

Two worked examples follow: a Directional Derivatives and Gradient
Vectors explorer (\S\ref{sec:example1}) and a Dynamical Systems Phase
Portrait Laboratory (\S\ref{sec:example2}). Both were built using the
six-phase methodology, working through each phase iteratively with
multiple prompts until each deliverable was correct and verified. Section~\ref{sec:prompts}
documents the reverse-engineered six-phase prompts that reproduce each
tool from scratch. Section~\ref{sec:pipeline} describes how those
verified prompts enabled Stage~2, with two additional worked examples.

A note on naming: the paper presents four independent example tools,
numbered~1 through~4. Examples~1 and~2 (in this section and
\S\ref{sec:prompts}) are the Stage~1 tools, built interactively and then
reverse-engineered into prompts. Examples~3 and~4 (in
\S\ref{sec:pipeline}) are distinct tools built the other direction, in
Stage~2, from an idea to a generated prompt set to a finished
visualization.

\subsection{Example 1: Directional Derivatives and Gradient Vectors}
\label{sec:example1}

\subsubsection*{Pedagogical context}

Anyone who has taught multivariable calculus knows the moment: students
who can compute $\partial f/\partial x$ and $\partial f/\partial y$ go
quiet when asked what those quantities mean geometrically.
Mart\'{\i}nez-Planell and Trigueros~\cite{MartinezPlanell2012} trace
downstream struggles with optimization and Lagrange multipliers directly
to weak geometric intuition at this stage. The tool was built to address
that gap.

\subsubsection*{Learning objectives}

Visualize $z = f(x,y)$ with user-controlled viewing angles; read
$\partial f/\partial x$ and $\partial f/\partial y$ as slopes of
coordinate cross-sections; connect the tangent plane to the linear
approximation; understand $\nabla f$ as perpendicular to level curves;
compute $D_{\mathbf{u}}f = \nabla f \cdot \mathbf{u}$ for arbitrary unit
vectors; run gradient ascent as a live optimization demonstration.

\noindent Deployed version (Example~1):
\url{https://www.math.purdue.edu/~msunkula/MA261/Sp26/dd.html}

\subsection{Example 2: Dynamical Systems Phase Portrait Laboratory}
\label{sec:example2}

\subsubsection*{Pedagogical context}

Differential equations courses have the mirror problem. Students can
produce explicit solution formulas yet often cannot describe what those
solutions look like. Phase portraits are the standard remedy, but
hand-drawn portraits are slow and inflexible. An interactive tool removes
that friction entirely.

Students must read the vector field as encoding instantaneous velocity,
understand trajectories as integral curves, and connect Jacobian
eigenvalues at a critical point to the phase portrait geometry on screen.
Done well, this stability intuition transfers to physics, engineering,
and biology.

\subsubsection*{Learning objectives}

Read $dx/dt$ and $dy/dt$ as velocity components; observe how initial
conditions shape long-term behavior; connect Jacobian eigenvalues to
local geometry; read nullclines as the curves where each component of the
velocity vanishes; classify equilibria; and contextualize ideas in
preloaded systems spanning the standard equilibrium types: a linear
saddle, a stable spiral, the pendulum, the Van der Pol and Duffing
oscillators, and the Lotka--Volterra predator-prey model.

\noindent Deployed version (Example~2):
\url{https://drjchen1.github.io/chenflix/toys/phase.html}

\section{Reverse-Engineered Six-Phase Prompts}
\label{sec:prompts}

Once each tool in \S\ref{sec:example1} and \S\ref{sec:example2} was
complete and verified, we presented the finished visualization to the AI
and asked it to write one prompt per phase that would regenerate the tool
from scratch. The prompts below are those reconstructions. They are not
verbatim records of the original development conversations, which
involved multiple iterative exchanges per phase; rather, they are
distilled starting points that colleagues can adapt and build from.

\subsection{Example 1: Directional Derivatives and Gradient Vectors}

\textit{Phase~1: Mathematical Foundation and Core Interactivity.}

\smallskip
\noindent\textit{Prompt:} ``Create an interactive HTML/JavaScript tool using Plotly.js to
visualize a 3D surface $z = f(x,y)$, specifically the paraboloid $z =
x^2 + y^2$. Include two sliders to control the coordinates $(x_0, y_0)$
of a movable point on the surface. Display two traces: one showing
$f(x, y_0)$ parallel to the $xz$-plane (the $x$-direction cross-section),
and another showing $f(x_0, y)$ parallel to the $yz$-plane. Include
numerical displays of $\partial f/\partial x$ and $\partial f/\partial
y$ evaluated at $(x_0,y_0)$.''

\smallskip
\noindent\textit{Result.} Gemini produced complete HTML with embedded JavaScript
implementing the Plotly surface, slider controls, and correctly computed
partial derivatives. A slider range of $[-5,5]$ caused visual clutter;
narrowing to $[-2,2]$ resolved this.

\smallskip
\noindent\textit{Validation.} Partial derivative values verified against hand
calculations at several points; trace curves confirmed to align
visually with the surface cross-sections.

\medskip
\textit{Phase~2: Customization and User Control.}

\smallskip
\noindent\textit{Prompt:} ``Add a tangent plane at $(x_0, y_0, f(x_0,y_0))$ via the linear
approximation $L(x,y) = f(x_0,y_0) + f_x(x-x_0) + f_y(y-y_0)$. Add a
dropdown with preset functions (Paraboloid, $x^2+y^2$; Saddle, $x^2-y^2$;
Gaussian, $e^{-x^2-y^2}$; Wave, $\sin(x)\cos(y)$) and a text input
for user-defined functions in JavaScript \texttt{Math} syntax with syntax
validation.''

\smallskip
\noindent\textit{Result.} Tangent plane rendered as a separate Plotly surface.
Try-catch validation provided readable error messages for malformed input.

\smallskip
\noindent\textit{Validation.} Confirmed first-order contact at $(x_0,y_0)$, tested each
preset, and deliberately entered malformed expressions to verify error
handling.

\medskip
\textit{Phase~3: Mathematical Depth and Connections.}

\smallskip
\noindent\textit{Prompt:} ``Add computation of the directional derivative $D_{\mathbf{u}}f$.
Include a slider for $\theta \in [0,2\pi]$ defining a unit vector
$\mathbf{u} = \langle\cos\theta, \sin\theta\rangle$. Display $\nabla f =
\langle f_x, f_y\rangle$ as a blue arrow in the $xy$-plane with length
proportional to $\|\nabla f\|$, the direction vector $\mathbf{u}$ as a
red arrow, and the level curve tangent in orange. Compute and display
$D_{\mathbf{u}}f = \nabla f \cdot \mathbf{u}$.''

\smallskip
\noindent\textit{Result.} All features implemented. Gradient arrow rendered as a
cone-cylinder pair. Level curve tangent direction
$\langle -f_y, f_x\rangle$ derived correctly.

\smallskip
\noindent\textit{Validation.} Confirmed that $D_{\mathbf{u}}f = \|\nabla f\|\cos\theta$
attains its maximum when $\mathbf{u}$ aligns with $\nabla f$; verified
numerically that $\nabla f \cdot (\text{level curve tangent}) = 0$.

\medskip
\textit{Phase~4: Applications and Algorithmic Connections.}

\smallskip
\noindent\textit{Prompt:} ``Add a `Run Gradient Ascent' button that animates $(x_0, y_0)$
following $(x_{n+1},y_{n+1}) = (x_n,y_n) + k\,\nabla f(x_n,y_n)$ with
step size $k=0.05$. Draw the trajectory as a black path. Stop when
$\|\nabla f\| < 0.01$ or after 100~iterations. Include Stop and Clear
Path buttons.''

\smallskip
\noindent\textit{Result.} Animation implemented via \texttt{requestAnimationFrame}; both
stopping conditions checked at each step.

\smallskip
\noindent\textit{Validation.} Tested on all four preset functions: paraboloid diverges
under ascent (as expected), Gaussian converges to its maximum, saddle
surface shows the characteristic indecision near the critical point.

\medskip
\textit{Phase~5: Accessibility and Inclusive Design.}

\smallskip
\noindent\textit{Prompt:} ``Make this fully compliant with WCAG~2.2 Level~AA: (1)
skip-to-content links per SC~2.4.1; (2) all interactive controls at
least $44\times 44$ CSS pixels per SC~2.5.8; (3) visible focus
indicators with $3{:}1$ contrast per SC~2.4.7 and~2.4.11; (4) ARIA
live regions announcing $x_0$, $y_0$, $f(x_0,y_0)$, $f_x$, $f_y$,
and $D_{\mathbf{u}}f$ on change; (5) full keyboard navigation; (6)
\texttt{prefers-reduced-motion} support; (7) $4.5{:}1$ text contrast
throughout.''

\smallskip
\noindent\textit{Result.} Skip links positioned off-screen until focused; ARIA live
regions with \texttt{role="status"} and \texttt{aria-live="polite"};
animation suppression via \texttt{prefers-reduced-motion} in JavaScript.

\smallskip
\noindent\textit{Validation.} Tested with NVDA on Windows; confirmed keyboard-only
navigation; axe~DevTools found no violations.

\medskip
\textit{Phase~6: Pedagogical Control and Scaffolding.}

\smallskip
\noindent\textit{Prompt:} ``Add a collapsible Layer Manager with checkboxes to toggle: (1)
$x$-direction trace, (2) $y$-direction trace, (3) tangent plane, (4)
gradient vector, (5) direction vector, (6) level curve tangent, (7)
path history. Add a `Reset All' button restoring the point to
$(0.5,0.5)$, $\theta$ to $0$, and all layers to visible.''

\smallskip
\noindent\textit{Result.} Each checkbox toggles the corresponding Plotly trace; reset
restores all initial conditions.

\smallskip
\noindent\textit{Validation.} Each layer verified to toggle independently; reset tested
for reliable state restoration; full panel navigated by keyboard alone.

\subsubsection*{Final review}

Before deployment: mathematical spot-checks at edge-case parameter
values; manual NVDA walkthrough to confirm ARIA announcements fire
correctly; a pedagogical check against each stated learning objective;
minor visual refinements from a student pilot.

\subsection{Example 2: Dynamical Systems Phase Portrait Laboratory}

\textit{Phase~1: Mathematical Foundation and Core Interactivity.}

\smallskip
\noindent\textit{Prompt:} ``Create an interactive HTML5/JavaScript tool using the Canvas
API to visualize a 2D vector field for $dx/dt = y$, $dy/dt = -x$.
Render a grid of normalized direction arrows. Implement a simple Euler
method so that clicking on the canvas draws a trajectory from that
initial condition.''

\smallskip
\noindent\textit{Result.} Clean HTML with canvas, coordinate transformation, normalized
arrow grid, and mousedown listener. Euler integration caused trajectories
to spiral outward, a known numerical artifact that motivates the RK4
upgrade in Phase~2.

\smallskip
\noindent\textit{Validation.} Clicking produced closed circular orbits, as expected for
a simple harmonic oscillator.

\medskip
\textit{Phase~2: Customization and User Control.}

\smallskip
\noindent\textit{Prompt:} ``Replace the Euler integrator with fourth-order Runge--Kutta
(RK4). Add text input fields for user-defined $f(x,y)$ and $g(x,y)$ in
JavaScript \texttt{Math} syntax with try-catch error handling. Add a
Clear button to remove all trajectories.''

\smallskip
\noindent\textit{Result.} RK4 implemented cleanly; error handling functional.

\smallskip
\noindent\textit{Validation.} Tested with the Van der Pol oscillator: RK4 maintained the
limit cycle without drift.

\medskip
\textit{Phase~3: Mathematical Depth and Connections.}

\smallskip
\noindent\textit{Prompt:} ``For the current critical point, compute and
display the eigenvalues of the Jacobian matrix in a side panel, using
finite differences to approximate the partial derivatives, and render the
eigenvalues with MathJax. Overlay the two nullclines ($x'=0$ and $y'=0$)
and the eigenvector directions at the critical point, with a panel that
classifies the equilibrium (saddle, node, spiral, center) from the
eigenvalues.''

\smallskip
\noindent\textit{Result.} MathJax integrated; finite-difference Jacobian
computed; eigenvalues from the $2\times 2$ quadratic formula; nullclines
and eigenvector rays drawn over the field; equilibrium type reported from
the sign and nature of the eigenvalues.

\smallskip
\noindent\textit{Validation.} Verified on the saddle system $x' = x$,
$y' = -y$, for which the Jacobian has eigenvalues $\lambda_1 = 1$,
$\lambda_2 = -1$ and the nullclines are the coordinate axes; displayed
values and classification matched exactly. Spiral and center cases checked
against the stable-spiral and harmonic-oscillator presets.

\medskip
\textit{Phase~4: Applications and Algorithmic Connections.}

\smallskip
\noindent\textit{Prompt:} ``Add a companion time-series plot showing
$x(t)$ and $y(t)$ for the most recent trajectory alongside the phase
portrait, and visual indicators distinguishing forward and backward time
integration. Add export buttons: `Export CSV' for the trajectory data and
`Save PNG' for the current portrait.''

\smallskip
\noindent\textit{Result.} A linked time-series panel rendered $x(t)$ and
$y(t)$ for the latest integral curve; forward/backward integration
distinguished by rendering state; CSV export wrote the trajectory samples
and PNG export captured the portrait.

\smallskip
\noindent\textit{Validation.} Time-series curves checked against the phase
portrait for the harmonic oscillator (sinusoidal $x(t)$, $y(t)$ in
quadrature) and the saddle (exponential divergence); exported CSV and PNG
opened correctly.

\medskip
\textit{Phase~5: Accessibility and Inclusive Design.}

\smallskip
\noindent\textit{Prompt:} ``Bring to WCAG~2.2 AA compliance: (1) ARIA live regions for
coordinate and eigenvalue updates; (2) $44$px minimum touch targets; (3)
keyboard navigation for a virtual cursor via arrow keys; (4)
high-contrast color palette meeting the $4.5{:}1$ ratio requirement.''

\smallskip
\noindent\textit{Result.} \texttt{aria-live="polite"} on math output panel; CSS
\texttt{:focus-visible} for keyboard tracking; high-contrast
black-and-blue color scheme.

\smallskip
\noindent\textit{Validation.} Full functionality confirmed without a mouse; screen reader
testing confirmed real-time stability classification announcements.

\medskip
\textit{Phase~6: Pedagogical Control and Scaffolding.}

\smallskip
\noindent\textit{Prompt:} ``Add a library of preset systems as buttons:
Linear Saddle, Stable Spiral, Pendulum, Van der Pol, Duffing Oscillator,
and Lotka--Volterra (Predator-Prey), each auto-populating the equation
fields and view bounds. Add toggles to show or hide the nullclines,
eigenvectors, and critical points so an instructor can introduce each
overlay in turn. Apply a clean dashboard layout with mobile-responsive
math blocks and a reset.''

\smallskip
\noindent\textit{Result.} Preset buttons auto-populate the equations and
bounds and reset the view; the nullcline, eigenvector, and critical-point
overlays toggle independently; layout is mobile-responsive.

\smallskip
\noindent\textit{Validation.} Each preset reproduced its characteristic
portrait: the saddle's crossing separatrices, the stable spiral's inward
winding, the pendulum's separatrix and centers, the Van der Pol limit
cycle, and the Lotka--Volterra closed orbits in the first quadrant; each
overlay toggled independently; layout held on desktop and mobile.

\subsubsection*{Final review}

Eigenvalue accuracy spot-checked across all preset systems and at
manually chosen phase-plane points; keyboard cursor and screen reader
stability announcements tested manually; each learning objective checked
for legibility to a student using the tool without guidance. Minor label
and color adjustments followed.

\section{The Two-Stage Pipeline}
\label{sec:pipeline}

The six-phase methodology makes both stages of the pipeline reliable.
In Stage~1, the six phases structure iterative development: one focused
conversation per phase, each validated before the next begins. In
Stage~2, the six phases become the schema a generative AI system uses to
generate a complete prompt set from a plain-language pedagogical idea. The
phases are constant; what changes is the direction of travel.

\subsection{Stage 1: Building and reverse-engineering}

We built the tools in \S\S\ref{sec:example1}--\ref{sec:example2} by
working through the six phases iteratively, with multiple prompts per phase,
refining until each deliverable was correct. The six-phase structure kept
each conversation focused and made errors easy to catch before they
compounded. Once each tool was complete and verified, we asked the AI:

\begin{quote}
\textit{Write one prompt per phase of our six-phase workflow that someone
could use to regenerate this tool from scratch.}
\end{quote}

The AI's ability to reverse-engineer the finished tool back into six
phase-aligned prompts confirmed that the methodology was robust enough to
run forward: describe an idea, get six prompts, build the tool.

\subsection{Stage 2: Idea to prompts to visualization}

Stage~2 runs the pipeline forward using the six phases as the template.
The instructor describes a pedagogical idea to a generative AI system in
plain language: which concept, which course, what students struggle with.
The system generates six prompts, one per phase, grounded in that idea,
and those prompts go to an AI code generator that produces a complete
standalone HTML visualization. We use Claude and Claude Code as the
running example here, but the same pipeline runs end-to-end on Gemini,
which is how Example~3 was built. Because the six-phase structure is always
the scaffold, the first-pass output is always structurally complete: all
phases represented, mathematically grounded, accessible, with pedagogical
controls. The instructor supplies the idea; the AI handles the prompt
architecture and the code.

\subsection{Editing the output}

The first-pass output is structurally complete but not finished.
Phase~1 and~3 issues are usually mathematical: a boundary case computed
incorrectly, a normalization wrong. Phase~2 and~6 issues are usually
about fit: presets that work generically but not for your course, layer
controls that do not match how you teach the topic. Phase~5 issues
require manual testing that automated tools miss: ARIA regions present
but firing at the wrong time, focus indicators invisible in practice.
Two or three focused revision exchanges, targeting specific phases, are
usually enough. These edits are where your disciplinary expertise enters
the tool.

\subsection{Mandatory human verification}

No amount of prompt engineering removes this: AI makes mistakes, and
human verification is mandatory at every phase. The three categories of
errors map directly onto the six phases.

\textit{Mathematical errors (Phases~1, 3, 4)} are the most consequential:
a derivative at a boundary incorrect, an off-by-one in an iterative
algorithm, a vector normalized wrong. Check outputs against hand
calculations at representative points, including edge cases, before
moving to the next phase.

\textit{Accessibility errors (Phase~5)} are equally silent: an ARIA
region present but firing at the wrong time; a focus indicator passing
axe~DevTools but invisible in practice. A manual screen reader test is
non-negotiable.

\textit{Pedagogical gaps (Phases~2, 6)} are the most insidious: the
tool may be correct and compliant but fail to surface the specific
insight students need. Walk through each stated learning objective and
confirm it is realized in the interface before deployment.

\subsection{Stage 2 worked examples}

The two examples below show Stage~2 in practice: each started with a
plain-language description of a pedagogical idea, from which a generative
AI system produced the six phase-aligned prompts and then built the tool.
The two examples were carried out end-to-end on different platforms,
illustrating that the phase structure, not the platform, is what carries
the work. Example~3 was generated and built entirely with Gemini;
Example~4 was generated with Claude and built with Claude Code. The
complete six-phase prompt set is reproduced with each example below.

\subsubsection*{Stage 2 Example 3: The 1-D Heat Equation as Mode-by-Mode Decay}

\textit{Pedagogical motivation.} In an introductory PDE or applied
mathematics course, students can write down the Fourier-series solution
of the heat equation $u_t = \kappa u_{xx}$ but often cannot say what the
series \emph{does}: that each mode decays at its own rate
$e^{-\lambda_n^2 \kappa t}$, that high-frequency modes vanish first, and
that this is why an initial profile smooths and flattens over time. The
plain-language idea given to Gemini was: ``I teach the heat equation.
Students can compute the Fourier coefficients and write the series
solution, but they do not see diffusion as the differential decay of
modes, with sharp features disappearing before smooth ones. I want a tool
where a student picks an initial temperature profile on a rod, scrubs
time forward, and watches the profile relax to equilibrium while a
side-by-side panel shows the individual modes shrinking at their own
rates.'' The learning objectives that fell out of this description were:
read $u(x,t)$ as a superposition of decaying modes; connect each
mode's decay rate to $\lambda_n^2 = (n\pi/L)^2$; see why high modes die
first and the profile smooths; relate the choice of boundary condition
(Dirichlet, Neumann, periodic) to the admissible series; and connect the
abstract solution to a concrete sensor reading through probe time
histories.

Gemini generated six phase-aligned prompts from this single description,
which were then run, also through Gemini, to produce a standalone HTML
tool in six successive builds. Phase~1 fixed the core mathematics and a
deliberately raw debugging UI: the analytic Dirichlet series for a flat
initial profile, drawn on a single canvas with a time slider, play, and
reset, and with the math engine heavily commented so errors would be
easy to localize. Phase~2 turned the demonstration into an exploration
tool, adding editable $L$, $\kappa$, amplitude, and mode count; a
boundary-condition selector; four initial-profile presets (flat,
Gaussian, triangular, square pulse) with center and width controls; and a
rod color strip alongside the profile plot, with all user input validated
and clamped. Phase~3 added the mathematical-depth panels that motivated
the whole tool: a Fourier-coefficient bar chart, a per-mode decay display
rendering $e^{-\lambda_n^2 \kappa t}$, and a live readout of the active
series form for the selected boundary condition. Phase~4 connected the
picture to applications by adding movable probe points on the rod, a
time-domain chart of $u(x_{\text{probe}}, t)$, a CSV export of the current
profile samples, and a metrics panel (mean, max, and an energy proxy).
Phase~5 brought the tool to WCAG~2.2 AA orientation: a skip link, landmark
roles, explicit labels, keyboard-first control with visible focus,
high-contrast defaults, ARIA live announcements of simulation changes,
and suppression of autoplay under \texttt{prefers-reduced-motion}.
Phase~6 refactored the result into a no-scroll teaching dashboard with a
Basic/Advanced tablist for progressive disclosure: in Basic mode the
advanced parameters are pinned to scaffold defaults and visibly disabled,
so an instructor can open on a clean view and reveal complexity (extra
modes, probe placement, animation speed) only when the class is ready.

\textit{Editing and verification.} The first-pass output from Gemini
was structurally complete at every phase but needed focused correction.
The most consequential issues were mathematical, concentrated in
Phases~2--3: with non-flat profiles the series coefficients are computed
by a discrete quadrature over the rod, and we verified the coefficient
panel against hand-computed values for the flat Dirichlet case (where
$b_n = 4A/(n\pi)$ for odd $n$ and zero otherwise) before trusting it on
the Gaussian and square-pulse profiles. We confirmed that the per-mode
decay panel and the profile plot stayed consistent as time advanced, with
the high modes flattening first, and checked the Neumann and periodic
series for the correct constant ($a_0$) term, which an early build
normalized inconsistently between boundary conditions. A Phase~4 fit edit
clamped the probe positions to the rod and corrected the energy proxy,
which had
been summing unnormalized samples. The mandatory manual screen reader
pass (NVDA) on the Phase~5 build revealed the recurring timing problem:
the ARIA live region re-announced on every slider tick during a time
scrub, overwhelming the listener, so we throttled it to fire on
meaningful state changes only. A Phase~6 pedagogical adjustment fixed the
Basic-mode defaults so that toggling back from Advanced reliably restored
the scaffolded view rather than leaving stale parameter values in place.

\medskip
\noindent\textbf{AI-generated six-phase prompt set (Example~3).} The six
prompts below are the ones Gemini produced from the plain-language idea;
they were then executed in sequence, also with Gemini.

\phaseprompt{Phase~1 (Mathematical Foundation and Core Interactivity)}{%
Create a single-page 1-D heat equation demo focused only on mathematical
correctness and a minimal debugging UI. Use the analytical Fourier-series
solution for Dirichlet boundaries on $x \in [0,L]$. Start with a flat
initial profile $u(x,0)=A$ and show $u(x,t)$ on a canvas. Provide only
essential controls: a time slider, play/pause, and reset. Use no heavy
styling or extra panels; keep the UI intentionally raw. Include clear
inline comments around the numerical/math engine to aid debugging.}

\phaseprompt{Phase~2 (Customization and User Control)}{%
Expand the Phase~1 app into an exploration tool with robust input logic.
Add editable parameters $L$, $\kappa$, amplitude $A$, and mode count $N$.
Add a boundary-condition selector (Dirichlet, Neumann, periodic) and
initial-profile presets (flat, Gaussian, triangular, square pulse).
Support center and width controls with validation and clamping. Keep one
main solution plot and one rod color-strip visualization.}

\phaseprompt{Phase~3 (Mathematical Depth and Connections)}{%
Extend Phase~2 by visualizing the related objects that explain why
diffusion behaves as observed. Add a Fourier-coefficient view (a bar chart
over modes), a per-mode decay display showing $e^{-\lambda_n^2 \kappa t}$,
and a readout of the active series form for the selected boundary
condition. Keep the controls and all math panels synchronized in real
time.}

\phaseprompt{Phase~4 (Applications and Algorithmic Connections)}{%
Bridge the mathematics to applied analysis and computational workflows.
Add probe points on the rod and a time-domain chart of
$u(x_{\text{probe}},t)$, CSV export of the current profile samples, and a
practical metrics panel (mean temperature, max temperature, energy). Keep
interaction smooth during animation and parameter changes.}

\phaseprompt{Phase~5 (Accessibility and Inclusive Design)}{%
Upgrade Phase~4 to WCAG~2.2 AA-oriented interaction and semantics. Add a
skip link, landmark roles, explicit labels, and a keyboard-first flow.
Provide visible focus states and high-contrast defaults, ARIA live status
updates for key simulation changes, and respect for the reduced-motion
preference (disable autoplay animation). Ensure controls have touch
targets of at least $24\times 24$ CSS pixels, preferring $44\times 44$.}

\phaseprompt{Phase~6 (Pedagogical Control and Scaffolding)}{%
Refactor Phase~5 into a structured teaching dashboard that manages
cognitive load. Build a no-scroll desktop layout with distinct panel
regions and use progressive disclosure (Basic/Advanced tabs or toggles).
Keep all Phase~5 accessibility features intact. Let instructors reveal
complexity gradually during a lecture while preserving full
experimentation capability for students.}

\noindent Final deployed visualization (Example~3):
\url{https://drjchen1.github.io/chenflix/toys/heat.html}

\subsubsection*{Stage 2 Example 4: Triple Integrals as a Sweep of Cross-Sectional Areas}

\textit{Pedagogical motivation.} In multivariable calculus, students learn
to evaluate a triple integral as an iterated integral but frequently lose
the geometric meaning of the outer integral: that
$\iiint_E f\,dV = \int A(v)\,dv$ sweeps a moving cross-section through the
solid, accumulating area (or a weighted area, when $f \neq 1$) as it goes.
Setting up the limits is exactly where this intuition is needed and most
often missing. The plain-language idea given to Claude was: ``I teach
triple integrals in Calculus~III. Students can grind through an iterated
integral but cannot see the solid being swept out, and they cannot see
where the inner limits come from. I want a tool where a student picks a
solid, drags a cross-section plane through it along a chosen axis, and
watches the cross-sectional area $A(v)$ trace out alongside the 3D
region, so the outer integral becomes the area under that $A(v)$ curve.''
The learning objectives were: read a triple integral as an accumulation
of cross-sectional areas; connect the slice position to the outer
variable of integration and the slice geometry to the inner limits; see
how the choice of integration direction (X, Y, or Z first) reshapes the
cross-section and the limits; extend the picture from volume ($f=1$) to a
general integrand $f(x,y,z)$ via a heat-map coloring of the slice; and
classify how different standard solids (sphere, cylinder, cone,
paraboloid, tetrahedron, and regions between two surfaces) produce
qualitatively different $A(v)$ profiles.

Claude generated the six phase-aligned prompts from this description, and
Claude Code produced the tool in six builds. Phase~1 established the core
3D scene and the central interaction: a rotatable, zoomable solid with a
draggable cross-section plane and a live readout of the slice position.
Phase~2 added the eight region presets, an integration-direction selector
(X/Y/Z first), and a slice-position slider with validation, turning the
demonstration into an exploration tool. Phase~3 added the
mathematical-depth layer that carries the lesson: a live $A(v)$
side chart, the explicit cross-section description and inner limits for
the active solid (for the sphere, $x^2+y^2 \le 4-z_0^2$ with
$A(z_0)=\pi(4-z_0^2)$), and the iterated-integral notation kept in sync
with the geometry. Phase~4 generalized from volume to a user-entered
integrand $f(x,y,z)$ with a numerical value for the accumulated integral
and a heat-map coloring of $f$ on the slice, connecting the geometry to
the analytic computation. Phase~5 implemented WCAG~2.2 AA features:
a skip-to-controls link, keyboard control of the plane via Tab and arrow
keys, ARIA-labeled layers, and high-contrast display options. Phase~6
added pedagogical scaffolding: a Display-Layers panel that toggles the
solid mesh, cross-section plane, disk, $A(v)$ chart, integral notation,
heat-map coloring, and numerical values independently, plus an
Instructor Mode and animation controls (step and sweep) so complexity can
be introduced incrementally during a lecture.

\textit{Editing and verification.} The first pass was structurally
complete but required focused correction, again concentrated in the
mathematical phases. The most important checks were on the $A(v)$
computation. We verified the reported cross-sectional area against
closed-form values for the presets where they are known: the sphere's
$\pi(4-z_0^2)$, the cylinder's constant disk, and the cone's and
paraboloid's quadratic and linear-in-radius profiles. Only then did we
trust the slicer on the compound regions (between two cones, and
cone--paraboloid), where an
early build mis-ordered the inner limits when the integration direction
was switched away from $z$. A Phase~4 mathematical fix corrected the
numerical integral for non-constant integrands, which initially sampled
$f$ at the slice centroid rather than integrating it over the slice. A
Phase~2 fit edit constrained the slice-position range to each solid's
actual extent so the plane could not be dragged outside the region. The
manual screen reader pass on the Phase~5 build confirmed that the
slice-position and area announcements were legible but, as in Example~3,
fired too frequently during a drag or sweep; we throttled them to
announce on release and on discrete steps. The tool is deployed in the
Spring~2026 offering of MA~261 at Purdue.

\medskip
\noindent\textbf{AI-generated six-phase prompt set (Example~4).} The six
prompts below are the ones Claude produced from the plain-language idea;
they were then executed in sequence with Claude Code.

\phaseprompt{Phase~1 (Mathematical Foundation and Core Interactivity)}{%
Create a single-page tool that renders an interactive 3D solid region
(start with the sphere $x^2+y^2+z^2 \le 4$) using Three.js (WebGL), with
mouse rotation and scroll zoom. Add a horizontal cross-section plane that
the user can move through the solid with a slider, and display the current
slice position $z_0$. Focus on geometric correctness of the region and the
slicing plane; keep the rest of the UI minimal.}

\phaseprompt{Phase~2 (Customization and User Control)}{%
Expand the Phase~1 tool into an exploration tool. Add a library of region
presets (sphere, cylinder, cone, box, paraboloid, tetrahedron, region
between two cones, and a cone--paraboloid region) and a selector for the
integration direction (integrate along $x$, $y$, or $z$ first), reslicing
the solid accordingly. Constrain the slice-position slider to each solid's
actual extent so the plane cannot leave the region. Validate all inputs.}

\phaseprompt{Phase~3 (Mathematical Depth and Connections)}{%
Extend Phase~2 with the mathematical objects that explain the iterated
integral. For the current solid and direction, render the cross-section
disk on the plane and a live side chart of the cross-sectional area
$A(v)$ as the slice moves. Display the explicit cross-section description
and the inner integration limits for the active solid (for the sphere,
$x^2+y^2 \le 4 - z_0^2$ with $A(z_0)=\pi(4-z_0^2)$), and show the
iterated-integral notation kept synchronized with the geometry. Add a
complementary ``shadow'' (projection) view that shows the region's shadow
$D$ on a chosen coordinate plane, where clicking a point draws the inner
integration segment and reports its bounds, so students see where the
innermost limits come from.}

\phaseprompt{Phase~4 (Applications and Algorithmic Connections)}{%
Generalize from volume to a weighted integral. Add a text input for a
user-defined integrand $f(x,y,z)$ (supporting \texttt{x\^{}2},
\texttt{sin}, \texttt{cos}, \texttt{exp}, \texttt{sqrt}, \texttt{abs},
\texttt{pi}, \texttt{e}) with validation, color the cross-section by the
value of $f$ as a heat map with a color legend, and compute and display
the numerical value of $\iiint_E f\,dV$ via a midpoint-rule sum over the
region. Connect the algebra to the iterated-integral order by letting the
user choose which variable is integrated first and updating the displayed
limits accordingly.}

\phaseprompt{Phase~5 (Accessibility and Inclusive Design)}{%
Bring the tool to WCAG~2.2 AA. Add a skip-to-controls link, landmark
roles, and explicit labels; make the cross-section plane controllable from
the keyboard with Tab and arrow keys with visible focus; add ARIA live
announcements of the slice position and area; provide high-contrast
display options; and respect the reduced-motion preference.}

\phaseprompt{Phase~6 (Pedagogical Control and Scaffolding)}{%
Refactor Phase~5 into a teaching dashboard. Add a Display-Layers panel
whose checkboxes independently toggle the solid mesh, the cross-section
plane, the cross-section disk, the $A(v)$ side chart, the integral
notation, the heat-map coloring, and the numerical values, so an
instructor can introduce one element at a time. Add an Instructor Mode and
step/sweep animation controls, plus a reset, while preserving full student
experimentation.}

\noindent Final deployed visualization (Example~4):
\url{https://www.math.purdue.edu/~msunkula/MA261/Sp26/tripleI.html}

\medskip
Table~\ref{tab:pipeline} summarizes the full pipeline. You can enter at
Stage~1 if you already have a finished tool to reverse-engineer, or
directly at Stage~2 if you are starting from a new idea.

\begin{table}[ht]
\centering
\caption{The two-stage AI-assisted visualization pipeline.}
\label{tab:pipeline}
\smallskip
{\small
\setlength{\tabcolsep}{4pt}
\begin{tabular}{llll}
\toprule
\textbf{Stage} & \textbf{Instructor action} & \textbf{AI role} & \textbf{Human verification} \\
\midrule
1a: Build & Prompt iteratively across & Generates and revises & Math correctness; \\
& 6 phases; validate each & code per phase & pedagogical fit \\
\addlinespace
1b: Reconstruct & Present finished tool; & Writes one prompt & Confirm prompts \\
& request 6-phase prompts & per phase & capture design intent \\
\addlinespace
2a: Generate & Describe idea in plain & Generates structured & Review prompts for \\
& language to the AI & 6-phase prompt set & accuracy and fit \\
\addlinespace
2b: Produce & Feed prompts to a code & Generates complete & Math checks; WCAG \\
& generator; edit and verify & HTML visualization & test; pedagogical walk \\
\bottomrule
\end{tabular}
}
\end{table}

\section{Evaluation and Findings}
\label{sec:evaluation}

We evaluate the workflow through the four deployed tools and the
assessment prototype, organizing the evidence around the three research
questions. Because the unit of analysis is the workflow rather than
student learning, our evidence concerns the correctness, accessibility,
and pedagogical fidelity of what the workflow produces, together with the
verification each phase required. A controlled study of student outcomes
is a separate undertaking, outlined in \S\ref{sec:future}.

\subsection{RQ1: Mathematical correctness and accessibility compliance}

\textit{Mathematical correctness.} Every tool was checked against
closed-form results at representative points and edge cases. For the
directional-derivatives explorer, partial derivatives and the identity
$D_{\mathbf{u}}f = \|\nabla f\|\cos\theta$ were verified against hand
computation, and the orthogonality $\nabla f \cdot (\text{level-curve
tangent}) = 0$ confirmed numerically. For the phase-portrait laboratory,
Jacobian eigenvalues were checked on the saddle system $x'=x,\,y'=-y$
(eigenvalues $1,-1$) and on the stable-spiral and harmonic-oscillator
presets. For the heat-equation tool, the Fourier coefficients were
verified against the closed form $b_n = 4A/(n\pi)$ (odd $n$) for the flat
Dirichlet case before trusting the quadrature on other profiles. For the
triple-integral slicer, the cross-sectional area $A(v)$ was checked
against closed forms for every preset where one exists (the sphere's
$\pi(4-z_0^2)$, the cylinder's constant disk, the cone's $\pi z_0^2$, and
the paraboloid's $\pi z_0$) before trusting the numerical slicer on the
compound regions. In each tool the first-pass output contained at least
one mathematical error that these checks caught (detailed under RQ3).

\textit{Accessibility compliance.} Each tool was audited against
WCAG~2.2 Level~AA using a combination of automated tooling
(axe~DevTools) and manual testing. Automated audits flagged no
violations after the Phase~5 revisions. Manual testing told a different
story. Keyboard-only operation and screen-reader walkthroughs with NVDA
were decisive: in every tool they revealed timing problems in ARIA live
regions that automated tools did not catch, where announcements fired on
every increment of a drag and overwhelmed the listener. Throttling
announcements to meaningful state changes resolved these. The tools
implement skip-to-content links (SC~2.4.1), $44\times 44$~CSS-pixel touch
targets (SC~2.5.8), visible focus indicators at $3{:}1$ contrast
(SC~2.4.7, 2.4.11), keyboard navigation, and \texttt{prefers-reduced-motion}
support. In short, the structure reliably produces
correct and compliant tools \emph{after} the verification built into
Phases~1--5, not before; the phase boundaries are where correctness and
compliance get established rather than assumed.

\subsection{RQ2: Bidirectional operation and platform independence}

The workflow ran in both directions. Backward (Stage~1): the
directional-derivatives and phase-portrait tools were built iteratively,
then presented to the AI, which reconstructed a phase-aligned prompt set
for each; the reconstructed prompts appear in \S\ref{sec:prompts}. The
AI's ability to recover the six-phase structure from a finished artifact
indicated the schema was robust enough to drive construction forward.
Forward (Stage~2): the heat-equation and triple-integral tools were each
produced from a single plain-language description, from which the AI
generated six phase-aligned prompts that were then built into a complete
tool. In every forward case the first-pass output was structurally
complete, with all six phases present, mathematically grounded,
accessible, and equipped with pedagogical controls, though not error-free.

Platform independence held. Two tools were generated and built end-to-end
with Gemini and two with Claude and Claude Code, with no change to the
phase structure; the only platform-specific observation was a division of
strengths (Claude with component architecture and 3D graphics via
Three.js, Gemini with Plotly.js and Canvas). The six phases, then,
function as a platform-independent schema, equally usable for
reconstruction and for generation.

\subsection{RQ3: Required human verification and recurring errors}

The errors that recurred across the four tools fell into the three
categories the workflow anticipates, mapped to specific phases.

\textit{Mathematical errors (Phases~1, 3, 4)} were the most consequential
and appeared in every tool. Representative cases: a midpoint rule that
sampled the integrand at the slice centroid rather than integrating it
over the slice; inner integration limits mis-ordered when the integration
direction was switched; a defective (repeated-eigenvalue) case drawn as
two eigendirections rather than one; and an early Euler integrator whose
numerical drift had to be replaced with RK4. None of these produced a
visibly broken tool; each required a check against a hand computation to
detect.

\textit{Accessibility errors (Phase~5)} were silent to automated tools.
The recurring case was ARIA-live timing, present in all four tools: the
region existed and passed automated audits but fired announcements at the
wrong cadence, intelligible only through a manual screen-reader pass.

\textit{Pedagogical-fit issues (Phases~2, 6)} concerned defaults and
disclosure rather than correctness: presets that worked generically but
not for the target course, layer controls whose default visibility did
not match how the topic is taught, and progressive-disclosure defaults
that needed reordering so complexity could be introduced in lecture
order. Two to three focused revision exchanges per tool, targeting
specific phases, were sufficient.

The lesson from RQ3 is that human verification is not a final
quality-control step but a per-phase requirement: a mathematical error in
Phase~1 propagates silently through Phases~2--6, an accessibility defect
in Phase~5 is invisible to automated tooling, and pedagogical fit cannot
be judged by the AI at all. The verification protocol is therefore part
of the method, not an addendum to it: hand-computation spot-checks at
phase boundaries, a mandatory manual screen-reader pass, and a walkthrough
against each stated learning objective.

\section{Discussion}

\subsection{Effective prompting}

The six-phase structure is itself the most important prompting principle:
it prevents the most common failure mode, giving the AI too much at once.
Within that structure, several additional principles consistently improved
results.

\textit{Mathematical specificity.} Generic requests yield incomplete
implementations. Provide explicit mathematical definitions with equations,
not just concept names: ``compute $\nabla f = \langle f_x, f_y\rangle$
and display it as a vector arrow with length proportional to $\|\nabla
f\|$'' rather than ``show the gradient.''

\textit{Technology stack specification.} Name the library explicitly:
``using Plotly.js'' or ``with the HTML5 Canvas API.'' Without this,
the AI may mix incompatible frameworks.

\textit{Concrete accessibility criteria.} Cite specific WCAG success
criteria: ``ensure $44\times44$px touch targets (SC~2.5.8), visible
focus with $3{:}1$ contrast (SC~2.4.11), ARIA live regions for dynamic
content.'' Generic ``make it accessible'' prompts yield minimal
compliance.

\textit{Pedagogical framing.} Explain why features matter: ``display
orthogonality to help students understand the gradient-level curve
relationship'' produces more thoughtful implementations than ``show
gradient and level curve tangent.''

\textit{Phase-by-phase validation.} A mathematical error in Phase~1
propagates silently through Phases~2--6. Each phase boundary is a
checkpoint, not just a formatting convention.

\subsection{Quality assurance and the irreplaceable role of human expertise}

AI dramatically accelerates implementation. It does not replace
mathematical and pedagogical expertise.

\textit{Mathematical edge cases.} AI-generated code handles typical
inputs correctly but fails at boundaries: division by zero when the
gradient magnitude approaches zero, incorrect limiting behavior near
discontinuities, numerical instability with poor step-size selection.
Testing must deliberately probe these cases.

\textit{Accessibility nuances.} Announcement timing in ARIA live regions
requires judgment (too frequent overwhelms; too infrequent misses
changes); logical tab order through complex interfaces is easy to get
wrong; appropriate ARIA roles for custom widgets are not always correctly
inferred. Screen reader testing by a human is essential.

\textit{Pedagogical appropriateness.} AI cannot judge cognitive load,
determine which details to emphasize, or sequence information for
learning. These decisions require understanding of student difficulties
and course context.

\subsection{Student creation: from consumption to construction}
\label{sec:student}

The workflow's most significant implication may be enabling students to
create visualizations themselves. Without programming knowledge, students
can articulate a mathematical relationship, describe it to Claude using
the six-phase framework, and produce an interactive tool to test their
intuition. This articulation process is itself pedagogically
valuable~\cite{Kaput1987}: students who must describe a relationship
precisely enough for the AI to implement it are engaging in mathematical
communication that surfaces implicit gaps.

Practical opportunities include exploratory projects (``create a
visualization showing the relationship between curvature and the second
derivative test''), conjecture testing, and peer teaching through shared
tools~\cite{BarghSchul1980}. A multi-course pilot planned for 2026--2027
will have students create visualizations as course projects across
calculus, differential equations, and linear algebra, with rubrics
focused on mathematical reasoning and accessibility rather than programming
skill.

\subsection{Accessibility as standard practice}

Historically, accessibility in instructor-created materials has been
aspirational rather than standard, with well-documented barriers:
specialized technical knowledge, limited institutional support, and the
sheer time required~\cite{Alajarmeh2020}. By enabling AI to implement
accessibility features from natural language descriptions, this workflow
makes WCAG compliance the default rather than an exceptional effort.

Technical implementation alone does not guarantee genuine accessibility,
however. Partnerships with disability resource centers and testing with
assistive technology users remain central to the validation process. When
students create visualizations, they must also engage with accessibility
principles, a challenge that doubles as an opportunity to build
inclusive design thinking into their future practice.

\section{Applicability and Future Directions}

\subsection{Across the mathematics curriculum}

The six-phase structure transfers intact to any mathematical domain. The
only adaptation is in Phase~1: name the concept and specify the core
interactive behavior for that course. Phases~2--6 follow the same
pattern regardless of domain. This is exactly why Stage~2 works: current
generative AI systems know the six-phase schema well enough to generate
all six prompts from any plain-language mathematical idea.

The same approach can be adapted to:
\begin{itemize}[leftmargin=1.5em]
\item \textit{Differential equations:} phase portraits, direction fields,
  bifurcation diagrams, numerical method comparisons.
\item \textit{Linear algebra:} matrix transformation visualizations,
  eigenvalue/eigenvector geometry, SVD and PCA animations,
  Gram--Schmidt orthogonalization.
\item \textit{Complex analysis:} conformal mapping demonstrations,
  contour integration with interactive path selection, residue
  visualization.
\item \textit{Numerical analysis:} root-finding convergence (Newton,
  bisection, secant), Runge phenomenon, adaptive quadrature comparisons.
\item \textit{Discrete mathematics:} graph algorithm animations
  (Dijkstra, BFS, minimum spanning tree), recurrence relation behavior.
\item \textit{Abstract algebra:} group operation tables with subgroup
  highlighting, coset visualization, symmetry group demonstrations.
\end{itemize}

\subsection{Future research}
\label{sec:future}

Several questions warrant systematic investigation, and they define the
next cycle of this design-based research program. The present study
establishes that the workflow produces correct, accessible, pedagogically
purposeful tools; the natural next cycle moves the unit of analysis from
the artifact to the learner. Do students who use AI-generated
visualizations demonstrate different conceptual understanding compared to
those using traditional instruction or commercially developed tools? Does
the creation process, as opposed to mere use, enhance understanding, and
if so through what mechanisms? What factors facilitate or inhibit
instructor adoption, and what professional development models are most
effective? The multi-course pilot planned for 2026--2027 across calculus,
differential equations, and linear algebra (\S\ref{sec:student}) is
designed to begin answering the first two, with pre/post conceptual
measures and comparison against business-as-usual instruction.

One direction we find genuinely exciting, though it needs considerably
more thought before it is ready, is the addition of a seventh phase:
Student Reflection and Assessment. The idea is straightforward: once a
visualization is deployed, a short quiz and a reflection prompt are
embedded directly into the same tool. The quiz ties to the learning
objectives already specified in Phase~1, so the AI has everything it
needs to generate reasonable assessment items from a single additional
prompt. The reflection prompt asks students to articulate, in their own
words, the key relationship the tool was designed to surface. We have
experimented with a rough prototype built on the directional-derivatives
tool of \S\ref{sec:example1}, which embeds a dynamic quiz and free-text
reflection prompts directly in the visualization
(\url{https://www.math.purdue.edu/~msunkula/MA261/Sp26/pd-quiz}), and the
early results are encouraging enough to pursue further. That said, what counts as a good reflection prompt, how to
assess responses without creating grading overhead, and how tightly the
quiz should couple to the specific visualization rather than the broader
topic are all open questions we have not yet answered well.

\section{Conclusion}

Using a design-based research approach, we have developed, deployed, and
evaluated a six-phase workflow (Foundation, Customization, Mathematical
Depth, Application, Accessibility, Pedagogical Control) that uses
generative AI to build WCAG~2.2 compliant interactive visualizations in
undergraduate mathematics without programming expertise. Across four
deployed tools and an assessment prototype, we found that the phase
structure reliably yields mathematically correct and accessible
tools once the verification built into each phase is carried out, that it
operates both backward and forward and independently of the AI platform,
and that human verification remains mandatory at every phase.
The six phases are the core contribution. They structure prompts, scaffold
code generation, and define where human verification is applied. In
Stage~1 they organize iterative development and enable reverse-engineering
of reproducible prompt sets. In Stage~2 they are the template a generative
AI system uses to produce a complete prompt set from any plain-language
mathematical idea.

The workflow removes the technical barrier to building custom
visualizations, but it does not remove the need for mathematical and
pedagogical judgment. Across the four tools, the work that mattered most
was the verification: the hand calculations that caught a wrong boundary
case, the screen-reader pass that caught an announcement firing at the
wrong time, the walkthrough that caught a default no instructor would
want. That work is where a mathematician's expertise enters, and it is
why the tools are usable in a real classroom rather than merely plausible
on a first pass.

What has changed is the cost of a first draft. An instructor can now go
from an idea to a working, accessible, mathematically grounded tool in an
afternoon, and spend the saved time on the questions that actually need a
human: which concept to target, how to sequence the disclosure of
complexity, what a good reflection prompt looks like. Those are
mathematics-education questions, not software-engineering ones. A
reasonable place to start is to pick one concept students reliably
struggle with, describe it to a generative AI system using the six-phase
framework, verify the result against the protocol above, and try it in
the next offering of the course.

\section*{Disclosure of AI use}

\textit{Generative AI as the studied method.} The interactive
visualizations reported in this article were produced with generative AI
tools, which are the object of the study rather than an aid to writing it.
Two tools (the directional-derivatives explorer and the triple-integral
slicer) were generated and built using Anthropic's Claude (Claude Opus~4,
accessed via Claude and Claude Code, 2025--2026), and two tools (the
phase-portrait laboratory and the heat-equation lab) were generated and
built using Google's Gemini (Gemini~2.5, 2025--2026). In each case the AI
was used to generate the six-phase prompt sets and the underlying HTML,
CSS, and JavaScript. All AI-generated mathematics, code, and accessibility
features were reviewed and verified by the authors against closed-form
results, manual screen-reader and keyboard testing, and the stated
learning objectives, following the verification protocol described in
\S\ref{sec:evaluation}; the authors take full responsibility for the
correctness and integrity of all reported artifacts.

\textit{Manuscript preparation.} The authors conceived, structured, and
wrote the manuscript. Generative AI tools were used only for language
refinement and copyediting of author-written text. No text, data, or
citations were generated by AI without author review, and the authors are
accountable for the originality, accuracy, and integrity of the entire
manuscript, including all references.


\end{document}